# DESENVOLVIMENTO DE UM BRAÇO MANIPULADOR ROBÓTICO: APLICAÇÃO DE INTERDISCIPLINARIDADE ENTRE PROJETO ASSISTIDO POR COMPUTADOR, MICROCONTROLADORES E ROBÓTICA INDÚSTRIAL

**Afonso Henriques Fontes Neto Segundo**
Mestre em Ciência da Computação pela Universidade Estadual do Ceará
Professor na Universidade de Fortaleza (UNIFOR)
Endereço: Av. Washington Soares, 1321 – Edison Queiroz, Fortaleza – CE, Brasil
E-mail: afonsof@unifor.br

**Joel Sotero da Cunha Neto**
Mestre em Informática Aplicada pela Universidade de Fortaleza
Professor na Universidade de Fortaleza (UNIFOR)
Endereço: Av. Washington Soares, 1321 – Edison Queiroz, Fortaleza – CE, Brasil
E-mail: joelsotero@unifor.br

**Reginaldo Florencio da Silva**
Mestrando em Ciência da Computação na Universidade Estadual do Ceará (UECE)
Pesquisador na Universidade de Fortaleza (UNIFOR)
Endereço: Av. Washington Soares, 1321 – Edison Queiroz, Fortaleza – CE, Brasil
E-mail: reginaldoflorencio@edu.unifor.br

**Paulo Cirillo Souza Barbosa**
Engenheiro de Controle e Automação pela Universidade de Fortaleza (UNIFOR)
Pesquisador na Universidade de Fortaleza (UNIFOR)
Endereço: Av. Washington Soares, 1321 – Edison Queiroz, Fortaleza – CE, Brasil
E-mail: pauloc@unifor.br

**Raul Fontenele Santana**
Graduando em Engenheiro de Controle e Automação pela Universidade de Fortaleza (UNIFOR)
Pesquisador na Universidade de Fortaleza (UNIFOR)
Endereço: Av. Washington Soares, 1321 – Edison Queiroz, Fortaleza – CE, Brasil
E-mail: raulfontenele@edu.unifor.br

***Resumo:*** *Este trabalho foi idealizado com base na Aprendizagem baseada em projetos (ABP) e apresenta as etapas de concepção, desenvolvimento e modelagem matemática de um manipulador robótico de baixo custo e com cinco graus de liberdade através de um projeto interdisciplinar, ligando duas disciplinas muito importantes do curso de Engenharia de Controle e Automação da Universidade de Fortaleza: Projeto Assistido por Computador, Microcontroladores e Robótica Indústrial. Ao fim são apresentados os resultados que o projeto trouxe para o melhor aprendizado da disciplina sobre a ótica do professor orientador e dos alunos.*
***Palavras-chave:*** *Robótica Industrial. Projeto Assistido por Computador. Microcontroladores. Cinemática. Robô Manipulador.*

# 1 INTRODUÇÃO

Um braço robótico é um dispositivo eletromecânico que consiste em articulações que podem ser acionadas por motores ou atuadores. Esses braços costumam ser guiado por sensores e controlados atrás de microcontroladores. Foram desenvolvidos com o objetivo de realizarem funções semelhantes ao braço humano e com grande precisão. Possuem diversas aplicações e são usados amplamente na indústria para a realização de tarefas repetitivas e que exigem grande precisão ou em situações perigosas (JAHNAVI, K.; SIVRAJ, P, 2017). Muitos manipuladores robóticos possuem um elevado custo devido a necessidade de atuadores de alta precisão e componentes personalizados e de difícil fabricação (TIO2, 2011).

A proposta deste trabalho é apresentar o desenvolvimento de um braço robótico de baixo custo com 5 graus de liberdade, que permita a integração entre as disciplinas de robótica industrial, microcontroladores e projeto assistido por computador, do curso de Engenharia de Controle e Automação da Universidade de Fortaleza (Unifor). A idealização da proposta foi inspirada na metodologia de aprendizado baseada em projetos (ABP), que segundo Barell (2010), é uma das mais eficazes formas disponíveis de envolver os alunos com o conteúdo que está sendo apresentado. A ABP é um formato de ensino no qual o aluno é motivado a aprender resolvendo problemas reais, que em muitos casos podem trazer contribuições às comunidades, o que implica em um maior envolvimento individual e uma maior interação com o grupo (BENDER, 2015).

# 2 DESENVOLVIMENTO DO MANIPULADOR ROBÓTICO

Para o desenvolvimento do manipulador, apelidado Jeri-Mun, foram inicial definidas algumas premissas para que braço pudesse ser usado não somente pelo alunos que o desenvolveram, mas também por todos os alunos da universidade que necessitarem realizar simulações em um braço robótico com 5 graus de liberdade em um ambiente laboratorial.

As premissas definidas para o Jeri-Mum foram:

1. Baixo custo em relação aos manipuladores disponíveis no mercado;
2. Possuir força suficiente para manipulação de objetos externos de até 1 kg;
3. Possuir movimentos precisos e suaves;
4. Conter a possibilidade receber coordenadas para a ponta do manipulador ou receber angulações para motores, ambas a distância, a fim de evitar o contato direto entre os alunos e braço;
5. Impedir a solicitação de movimentações inválidas a serem realizadas pelo sistema. Dessa forma, evitando possíveis colisões;
6. Ser fabricado com peças de fácil reposição.

Partindo destas premissas optou-se pela utilização de impressão 3D para fabricação das partes mecânicas do manipulador uma vez que a universidade possui uma impressora 3D para a utilização em projetos disciplinares. Para o controlador, foi escolhido a plataforma de prototipagem Arduino, por ser uma plataforma aberta e de fácil manipulação, enquanto para os atuadores foram escolhidos os servos motores MG995, uma vez que os mesmos possuem um torque de 8,5 kgf*cm quando alimentados em 5V. (Colocar o datasheet como fonte??)

### 2.1 Confecção e montagem do manipulador

O manipulador, nomeado Jeri-Mum, é formado por um conjunto de corpos conectados em cadeia por juntas, esses corpos são chamados de elos. A fim de imprimir todas as peças do manipulador, foi utilizado o software CATIA V5 para fazer a modelagem 3D do braço, elo por elo, ilustrado na Figura 1.

Figura 1. Montagem do Jeri-Mun no CATIA.

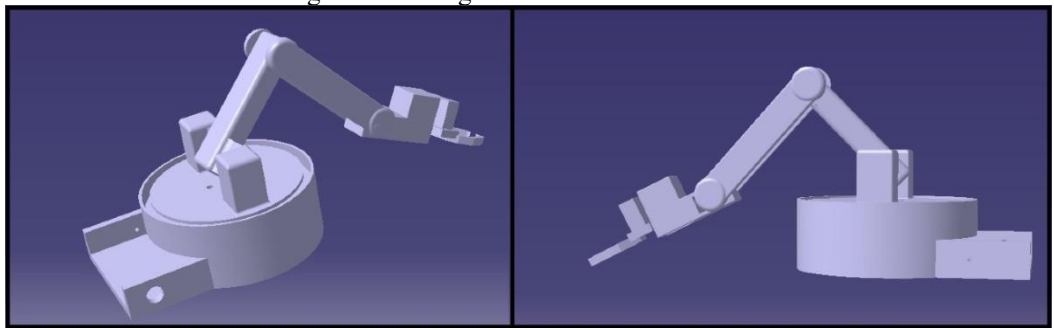

Fonte: O autor

Na Figura 2, pode-se ver o manipulador finalizado, juntamente com um rótulo indicando cada um dos seus cinco eixos, portanto, pode-se afirmar que o braço possui 5 graus de liberdade. Para uma maior noção de tamanho do mesmo, a distância entre os eixos 3 e 4 é de 170 milímetros.

Figura 2. Jeri-Mun finalizado.

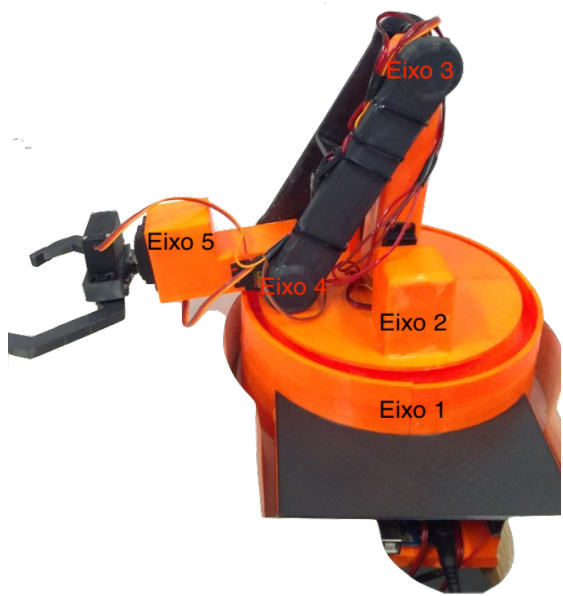

Fonte: O Autor

### 2.2 Cinemática

Na ponta de um manipulador robótico encontra-se a ferramenta necessária para execução da tarefa para a qual o manipular foi desenvolvido. Para o projeto em questão, a ferramenta utilizada foi uma garra, que pode ser programada para executar uma série de movimentações que a leve para um conjunto de posições e orientações de um plano cartesiano base. Uma das grandes contribuições deste trabalho é o desenvolvimento matemático das cinemáticas direta e inversa que possibilitam o planejamento da trajetória da ferramenta.

A cinemática direta permite determinar a posição e orientação da ferramenta em função das variáveis das juntas do manipulador em relação a sua base. Ou seja, é possível obter as coordenadas (x, y, z) da garra, utilizando as angulações de cada junta e o comprimento de cada elo. Enquanto a cinemática inversa faz exatamente o contrário da cinemática direta, ou seja, a partir da posição desejada em coordenadas (x, y, z) obtêm os ângulos necessários para o robô alcançar essas coordenadas, consistindo em descobrir as angulações de cada junção (COCOTA, 2013).

### 2.1.1 Cinemática Direta

Para realizar a cinemática direta, a convenção de Denavit-Hartenberg foi utilizada, já que a mesma permite a obtenção da posição e da orientação da ferramenta e estabelece, dentre outras coisas, que o comprimento e a torção de um elo qualquer dependem das juntas adjacentes. Dessa forma o sistema fica totalmente interdependente e a movimentação de qualquer junta impacta na posição da ferramenta (COCOTA, 2013).

Para o estudo e orientação de um corpo é necessário usar um sistema cartesiano como referência. Assim tem-se um sistema cartesiano fixo e um outro em relação ao qual o corpo irá se movimentar. Para representar um sistema de coordenadas em relação a outro, são usadas as matrizes de rotação, que irão representar a rotação do sistema em relação ao eixos x, y ou z (livrinho da unifor, alguma ano).

De acordo com Carvalho et al (2017), as matrizes homogêneas de rotação são:

Para o eixo x:

$$\begin{pmatrix} 1 & 0 & 0 & 0 \\ 0 & \cos(\theta) & -\sin(\theta) & 0 \\ 0 & \sin(\theta) & \cos(\theta) & 0 \\ 0 & 0 & 0 & 1 \end{pmatrix}$$

Para o eixo y:

$$\begin{pmatrix} \cos(\theta) & 0 & \sin(\theta) & 0 \\ 0 & 1 & 0 & 0 \\ -\sin(\theta) & 0 & \cos(\theta) & 0 \\ 0 & 0 & 0 & 1 \end{pmatrix}$$

Para o eixo z:

$$\begin{pmatrix} \cos(\theta) & -\sin(\theta) & 0 & 0 \\ \sin(\theta) & \cos(\theta) & 0 & 0 \\ 0 & 0 & 1 & 0 \\ 0 & 0 & 0 & 1 \end{pmatrix}$$

No caso das matrizes de translação, essas mostram o deslocamento feito pelo elo sobre o eixo x, y ou z. Sua representação de acordo com FOLEY et al. (1982):

$$\begin{pmatrix} 1 & 0 & 0 & \Delta_x \\ 0 & 1 & 0 & \Delta_y \\ 0 & 0 & 1 & \Delta_z \\ 0 & 0 & 0 & 1 \end{pmatrix}$$

Sendo os valores de x, y e de z referentes a translação realizada.

A Figura 3 mostra um esquemático do braço visto lateralmente em sua posição inicial de repouso. Para a realização dos cálculos da cinemática direta as distâncias entre as juntas do braço robótico são as seguintes: a1 = 63mm para a base, e para os demais eixos: a2 = 145mm; a3 = 170mm e a4 = 110mm.

Figura 3 – Representação esquemática do braço em repouso

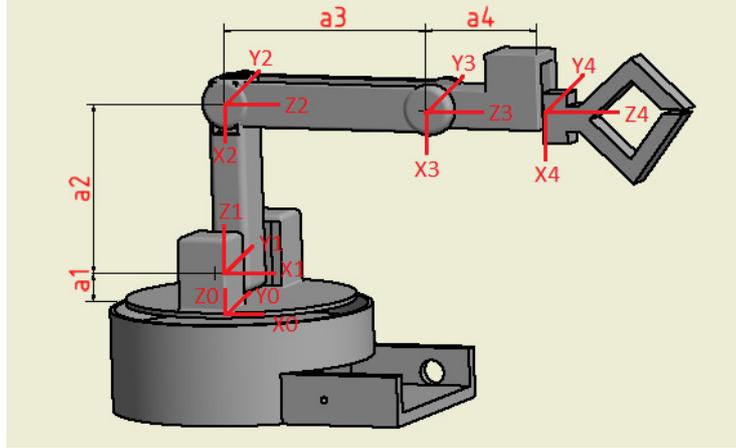

Fonte: Os autores

Dessa forma, para cada elo de ligação haverá uma matriz equivalente que será formada pela multiplicação das respectivas matrizes de rotação e translação. Neste projeto, as matrizes que representam cada um desses elos são representadas pelas seguintes equações:

$$M_1^0 = \begin{pmatrix} \cos(\theta_1) & -\sin(\theta_1) & 0 & 0 \\ \sin(\theta_1) & \cos(\theta_1) & 0 & 0 \\ 0 & 0 & 1 & a_1 \\ 0 & 0 & 0 & 1 \end{pmatrix} x \begin{pmatrix} \cos(\theta_2) & 0 & \sin(\theta_2) & 0 \\ 0 & 1 & 0 & 0 \\ -\sin(\theta_2) & 0 & \cos(\theta_2) & 0 \\ 0 & 0 & 0 & 1 \end{pmatrix}$$

$$M_2^1 = \begin{pmatrix} 1 & 0 & 0 & 0 \\ 0 & 1 & 0 & 0 \\ 0 & 0 & 1 & a_2 \\ 0 & 0 & 0 & 1 \end{pmatrix} x \begin{pmatrix} \cos(90+\theta_3) & 0 & \sin(90+\theta_3) & 0 \\ 0 & 1 & 0 & 0 \\ -\sin(90+\theta_3) & 0 & \cos(90+\theta_3) & 0 \\ 0 & 0 & 0 & 1 \end{pmatrix}$$

$$M_3^2 = \begin{pmatrix} 1 & 0 & 0 & 0 \\ 0 & 1 & 0 & 0 \\ 0 & 0 & 1 & a_3 \\ 0 & 0 & 0 & 1 \end{pmatrix} x \begin{pmatrix} \cos(\theta_4) & 0 & \sin(\theta_4) & 0 \\ 0 & 1 & 0 & 0 \\ -\sin(\theta_4) & 0 & \cos(\theta_4) & 0 \\ 0 & 0 & 0 & 1 \end{pmatrix}$$

$$M_4^3 = \begin{pmatrix} 1 & 0 & 0 & 0 \\ 0 & 1 & 0 & 0 \\ 0 & 0 & 1 & a_4 \\ 0 & 0 & 0 & 1 \end{pmatrix} x \begin{pmatrix} \cos(\theta_5) & -\sin(\theta_5) & 0 & 0 \\ \sin(\theta_5) & \cos(\theta_5) & 0 & 0 \\ 0 & 0 & 1 & 0 \\ 0 & 0 & 0 & 1 \end{pmatrix}$$

No caso da matriz $M_1^0$, esta representa o elo que liga as juntas 0 e 1, sendo o resultado de uma matriz transladada e rotacionada no eixo z de a1 e de θ1 respectivamente com uma outra matriz rotacionada θ2 em y. Já a matriz $M_2^1$ representa o elo de ligação das juntas 1 e 2, conforme a Figura 2, é o resultado de uma matriz transladada a2 em z em conjunto com uma matriz rotacionada 90 graus + θ3 no eixo y, nesse caso isso se faz necessário devido a diferença em relação a posição de repouso do braço conforme é mostrado na Figura 1. Já a matriz $M_3^2$ representa o elo de ligação das juntas 2 e 3, sendo o resultado de uma translação no eixo z de a3 acompanhada de uma rotação em y de θ4. Por último a matriz $M_4^3$ demonstra o último elo de ligação, da junta 3 a 4, sendo esta última a garra do robô, essa matriz é o resultado de uma translação de a4 em z com uma rotação de θ5 também em z.

Ao final, será calculada a matriz equivalente formada pela multiplicação de todas as anteriores da seguinte forma:

$$M_4^0 = M_1^0 x M_2^1 x M_3^2 x M_4^3$$

Assim, a partir da matriz representada pela Equação (5) será possível mostrar, com base nas rotações e translações ocorridas ao longo da estrutura do braço, na primeira, segunda e terceira linha da última coluna as coordenadas x, y e z, respectivamente. Determinando-se assim a posição e a orientação do efetuador.

### 2.1.2 Cinemática Inversa

Como visto anteriormente, através da cinemática direta é possível encontrar a posição final de uma ferramenta a partir de seus ângulos de junta bem como as possíveis translações em elos. Ao passo que na cinemática inversa, a partir das coordenadas da posição pretendida e da orientação, é possível encontrar os valores dos ângulos e transações que necessitam ser realizados para que a ponta da ferramenta se desloque até a coordenada (CRAIG, 2012). Para Craig, a grande dificuldade em realizar a modelagem através da cinemática inversa está relacionado ao fato de que o método de resolução de suas equações é de forma não linearidade. Este método tem como características seu alto custo de interação, poucos pontos de interpolação e garantia de chegada ao final da trajetória, mesmo passando por uma região não alcançável (ROCHA, 2017).

Para o manipulador desenvolvido neste projeto, o desafio está em encontrar os ângulos referentes a cada um dos três eixos que posicione a garra em um conjunto de coordenadas globais fornecidas previamente, levando em consideração que o elo representado por a4 sempre estará paralelo com o solo. Sendo assim, é necessário o uso de das vistas superior e lateral para obtenção das equações de cinemática inversa.

Considerando o braço em repouso como visualizado na Figura 2, e sua vista superior ilustrada pela Figura 3, é possível identificar somente os valores de x e y uma vez que desta vista não é possível visualizar o valor de z. A partir destes valores, podemos traçar uma linha que equivalente a projeção que a1,a2 e a3 possam fazer somando-se com a medida a4 que a mesma é paralela a mesa onde o braço se encontra, completando assim um triângulo retângulo a qual é denominada de ω+a4 e que pode ser calculado através da equação de pitágoras em 1.1. Para obtenção do ângulo $θ1$, iremos utilizar a relação trigonométrica mostrada na equação 1.2.

$$w = \sqrt{x^2 + y^2} - a4 \qquad (1.1)$$

$$θ1 = atan\left(\frac{y}{x}\right) \qquad (1.2)$$

Figura 4. Vista superior do braço manipulador Jeri-Mun.

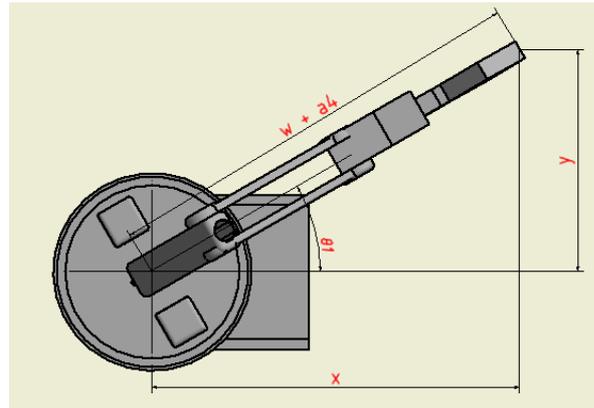
Fonte: O Autor

Como não há outro ângulo a ser calculado na vista superior, podemos prosseguir para vista lateral que é ilustrada pela Figura 5, para assim calcularmos os valores dos ângulos *θ2* e *θ3*. Traçando uma linha de *θ2* até *θ4*, representada na Figura 5 por $k$, é possível formar dois triângulos, um escaleno e um retângulo. Com as medidas ω e (z-a1) podemos calcular o valor de k utilizando o teorema de pitágoras representado pela Equação 1.3. Deste momento temos os valores de todas as medidas dos dois triângulos, logo podemos calcular o valor de *θ2* e *θ3* pelas Equações 1.4 e 1.5 respectivamente bastando somente encontrar os valores de $\alpha$, $\gamma$ e $\beta$, os quais são encontrados utilizando as equações 1.6, 1.7 e 1.8 respectivamente.

Figura 5. Vista lateral do braço

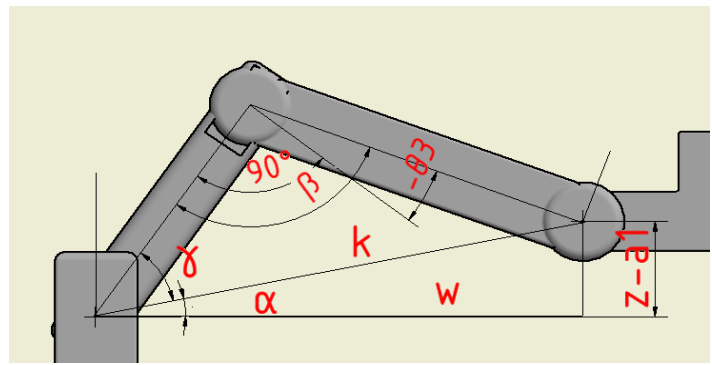
Fonte: O Autor

$$k = \sqrt{w^2 + (z-a1)^2} \qquad (1.3)$$

$$\theta 2 = 90 - \alpha - \gamma \qquad (1.4)$$

$$\theta 3 = 90 - \beta \qquad (1.5)$$

$$\alpha = atan\left(\frac{z-a1}{w}\right) \qquad (1.6)$$

$$\gamma = acos\left(\frac{a2^2 + k^2 - a3^2}{2*a2*k}\right) \qquad (1.7)$$

$$\beta = acos\left(\frac{a2^2 + a3^2 - k^2}{2*a2*a3}\right) \qquad (1.8)$$

Já para o ângulo $\theta 4$, que está associado ao eixo responsável por manter a garra do manipulador sempre em paralelo à mesa onde o mesmo opera, é intuitivo dizer que o mesmo deve desfazer a angulação dos valores de $\theta 2$ e $\theta 3$, como expressa a Equação 1.9.

$$\theta 3 = 90 - \beta \theta 4 = -\theta 2 - \theta 3 \qquad (1.9)$$

Não se faz necessário o cálculo do ângulo $\theta 5$ já que o mesmo é responsável somente pela abertura e fechamento da garra, e isto depende do que se deseja fazer com o manipulador.

### 2.1.3 Validação dos cálculos

O processo de validação das cinemáticas se dá em fornecer ângulos arbitrários e encontrar as coordenadas da ponta do manipulador através da cinemática direta, e fornecer estes valores para a cinemática inversa, obtendo os ângulos de entrada. Como, para um manipulador com cinco graus de liberdade, existe mais de uma combinação de ângulos possíveis que cheguem as mesmas coordenadas, se faz necessário realimentar a cinemática direta com os ângulos obtidos pelo cálculo. Pode-se então afirmar que os cálculos estão corretos se, ao final deste novo ciclo de cálculos pela cinemática direta, for obtido os mesmos valores de x, y e z fornecidos como entrada. As modelagens do braço foram então consideradas válidas.

## 3  INTERFACE PARA PRÁTICAS E PROJETOS FUTUROS

Para que o manipulador robótico possa ser utilizado como bancada experimental e/ou como parte de um experimento interdisciplinar que englobe mais disciplinas, foi desenvolvido uma interface de envio, recepção e validação de dados através de uma comunicação *bluetooth* mestre-escravo entre duas placas de prototipagem microcontroladas. O escravo está ligado na base do braço e o mestre não está fixo em nenhum local já que este é utilizado para enviar os comandos para o manipulador de forma remota. As informações são transmitidas sempre de forma unidirecional, do mestre para o escravo e podem conter tanto dados dos ângulos desejados para cada eixo como coordenadas onde deseja-se posicionar a ponta da ferramenta do manipulador robótico.

A Figura 3 mostra a bancada mestre com o módulo *bluetooth*. Nela feita a transmição de informação somente quando há alteração nos parâmetros. Estes dados podem ser adquiridos de 3 maneiras: recebendo as informações das angulações de cada eixo pela serial, recebendo as coordenadas absolutas (x,y,z) de onde deseja-se posicionar a ponta da ferramenta pela serial ou convertendo as angulações de cada potênciômetro em um ângulo desejado para cada eixo.

Os tipos de dados são identificados e tratados antes de serem transmitidos ao módulo escravo. Caso o dado informado seja um conjunto de ângulos desejados para cada eixo, faz-se a conversão dos mesmos para radianos. Caso seja informado um conjunto de coordenadas desejadas será necessario calculara cinemática inversa do manipulador. Sendo assim, os dados tratados sempre resultam em um conjunto de ângulos em radiano. Essa lógica está ilustrada no fluxograma da Figura 7.

Figura 6. Arduino mestre com controle via analógicas

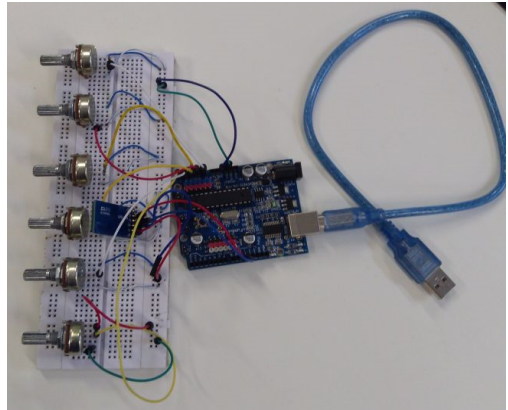

Fonte: O autor

Figura 7. Fluxograma das informações no Arduino mestre.

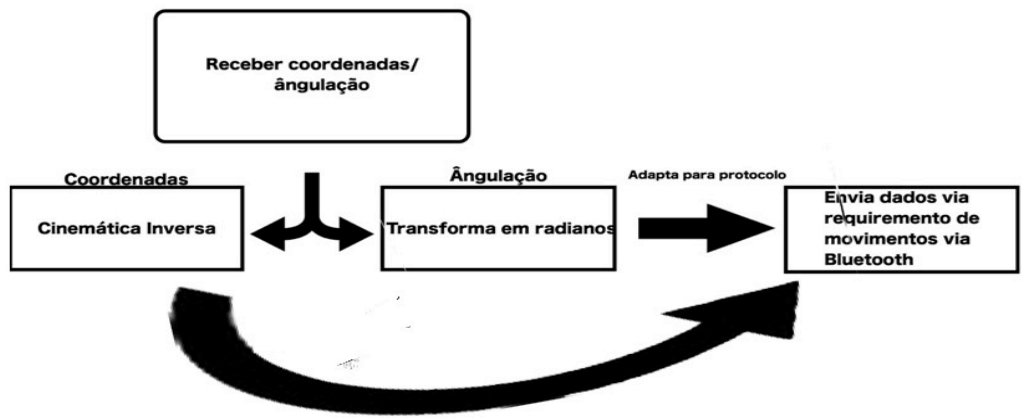

Fonte: O Autor

O microcontrolador escravo, que está conectado diretamente ao manipulador robótico, recebe as informações por bluetooth e aplica a cinemática direta nos ângulos para verificar se é uma coordenada válida e, se for, envia os ângulos para os servo motores. Essa lógica está ilustrada na Figura 8. Será dado como coordenada válida aquela que não leve o manipulador a colidir com ele mesmo nem com a bancada onde foi instalado.

Figura 8. Fluxograma das informações no Arduino escravo.

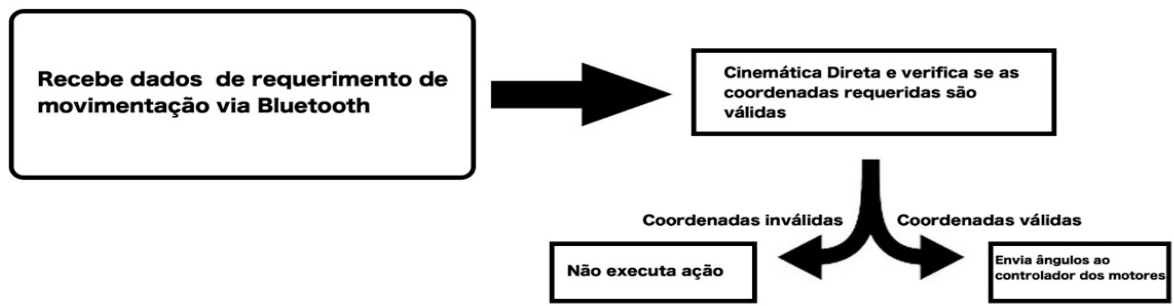

Fonte: O autor

## CONSIDERAÇÕES FINAIS

A utilização da metodologia de aprendizagem através de projetos mostrou resultados bastante satisfatórios na disciplina, tanto com o aumento do interesse dos alunos, por estarem desenvolvendo algo que possui aplicações reais por conta própria, quanto no aprendizado do conteúdo. Os alunos também se sentiram motivados por estarem no controle da situação e não apenas como ouvintes em uma sala de aula, assim estavam sempre buscando conhecimento e atrás dos professores para tirarem dúvidas para o aprimoramento do projeto.

Após a conclusão do trabalho foi realizado uma pesquisa com os alunos envolvidos no desenvolvimento do projeto com o propósito de avaliar a percepção dos mesmos sobre o trabalho desenvolvido. Foi pedido para que os mesmos atribuíssem notas de 1 a 5, onde 1 representava muito pouco ou nada e 5 representava muito, para o quanto o desenvolvimento do braço robótico contribuiu para o seu aprendizado da matéria e o como a turma estava engajada para a conclusão do trabalho. Cerca de 82% dos entrevistados atribuíram nota 4 ou 5 para a contribuição do aprendizado e para o engajamento da turma, porém houve um maior percentual de notas 5 para a avalição pessoal de desempenho.

Durante a aplicação do projeto foi possível notar o desenvolvimento de várias habilidades interpessoais como, por exemplo, espírito de equipe, proatividade, liderança, capacidade de resolver problemas, capacidade de tomada de decisões e gestão de projetos. Sendo assim. que os alunos envolvidos resgataram diversos conhecimentos vistos também em outras disciplinas da graduação e que são requisitos para o mercado de trabalho.

Por fim, para todos os engajados, o projeto pôde ter uma satisfatória experiência, tanto para os alunos quanto para os professores e provando, sim, que esta metodologia mostra-se um método bom no processo de ensino ou aprendizagem dos alunos, podendo ainda ser aplicado em outros projetos interdisciplinares.

**REFERÊNCIAS**

# DEVELOPMENT OF A ROBOTIC MANIPULATOR: APPLICATION OF INTERDISCIPLINARITY BETWEEN COMPUTER ASSISTED PROJECT, MICROCONTROLLERS AND INDUSTRIAL ROBOTICS

***Abstract:*** *This work was conceived based on Project-Based Learning (ABP) and presents the design, development and mathematical modeling steps of a low-cost robotic manipulator with five degrees of freedom through an interdisciplinary project linking two very important disciplines of the course of Control Engineering and Automation of the University of Fortaleza: Computer Aided Design, Microcontrollers and Industrial Robotics. At the end are presented the results that the project has brought to the best learning of the discipline on the optics of the tutor and students.*

***Key-words****: Industrial Robotics. Computer Assisted Design. Microcontrollers. Kinematics. Robot Manipulator.*